\documentclass[11pt,a4paper]{article}
\usepackage[hyperref]{emnlp2020}
\usepackage{times}
\usepackage{latexsym}
\usepackage{graphicx}

\usepackage{soul}
\usepackage{xcolor}
\usepackage{colortbl}
\sethlcolor{yellow}
\usepackage{mathtools}
\hypersetup{
    colorlinks = true
    }

\definecolor{orange}{rgb}{1,0.5,0}
\definecolor{lightsalmonpink}{rgb}{1.0, 0.6, 0.6}
\definecolor{verylightsalmonpink}{rgb}{0.966, 0.805, 0.797}
\definecolor{lightblue}{rgb}{0.862, 0.906, 0.984}
\definecolor{lightyellow}{rgb}{1.0, 0.945, 0.797}
\definecolor{lightgreen}{rgb}{0.835, 0.91, 0.828}
\definecolor{lightpurple}{rgb}{0.879, 0.832, 0.902}
\definecolor{green}{rgb}{0, 0.5, 0}
\usepackage{pifont}
\newcommand{\cmark}{\ding{51}}%
\newcommand{\xmark}{\ding{55}}%

\usepackage{microtype}

\aclfinalcopy 

\newcommand{\subq}[0]{\emph{sub-question}}
\newcommand{\irrq}[0]{\emph{irrelevant question}}
\newcommand{\subqs}[0]{\emph{sub-questions}}
\newcommand{\irrqs}[0]{\emph{irrelevant questions}}

\title{SOrT-ing VQA Models : Contrastive Gradient Learning \newline for Improved Consistency}

\author{Sameer Dharur$^1$ \quad Purva Tendulkar$^{1\rightarrow3}$ \\[0.05in]
  \textbf{Dhruv Batra$^{1,2}$ \quad Devi Parikh$^{1,2}$ \quad Ramprasaath R. Selvaraju$^{1\rightarrow4}$} \\[0.1in]
  {\tt\small \{sameerdharur, purva, dbatra, parikh, ramprs\}@gatech.edu}\\ \\[0.1in]
  $^1$Georgia Tech \quad
  $^2$Facebook AI Research \quad
  $^3$University of California, San Diego \quad
  $^4$Salesforce Research}

\date{}

\begin{document}
\maketitle

\begin{abstract}

Recent research in Visual Question Answering (VQA) has revealed state-of-the-art models to be inconsistent in their understanding of the world -- they answer seemingly difficult questions requiring reasoning correctly but get simpler associated sub-questions wrong. 
These sub-questions pertain to lower level visual concepts in the image that models ideally should understand to be able to answer the higher level question correctly. To address this, we first present a gradient-based interpretability approach to determine the questions most strongly correlated with the reasoning question on an image, and use this to evaluate VQA models on their ability to identify the relevant sub-questions needed to answer a reasoning question. 
Next, we propose a contrastive gradient learning based approach called Sub-question Oriented Tuning (SOrT) which encourages models to rank relevant sub-questions higher than irrelevant questions for an $<$image, reasoning-question$>$ pair. 
We show that SOrT improves model consistency by upto 6.5\% points over existing baselines, while also improving visual grounding. 
\end{abstract}
\section{Introduction}

\begin{figure*}[t!]
\includegraphics[scale=0.5]{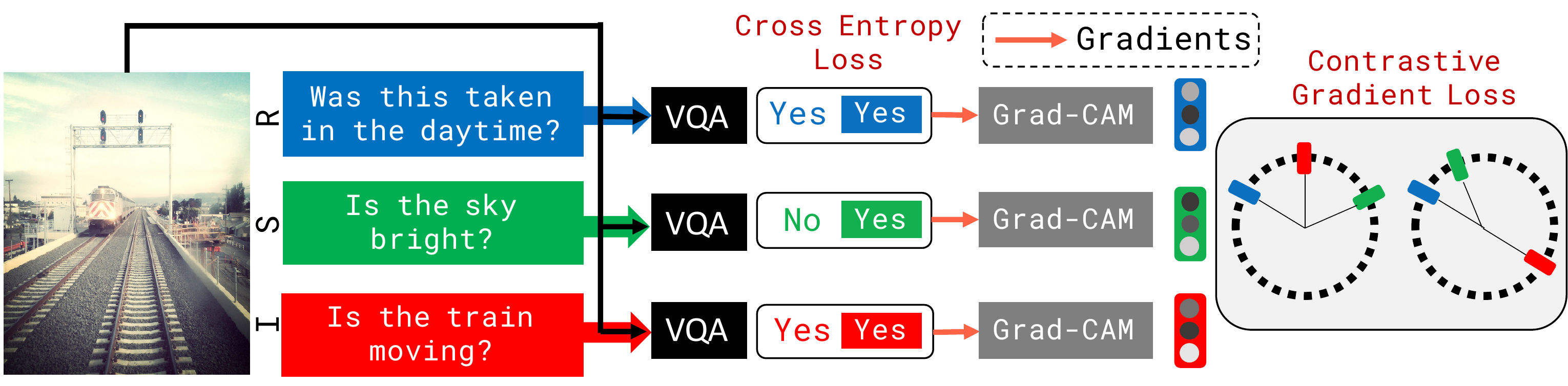}
\caption{The approach for SOrT. The reasoning question \textit{Was this taken in the daytime?} has the sub-question \textit{Is the sky bright?} and an irrelevant question \textit{Is the train moving?} We tune the model using cross entropy losses and a contrastive gradient loss to bring the reasoning question's Grad-CAM vector closer to that of its sub-question and take it farther away from that of an irrelevant question.}
\label{fig:approach}
\end{figure*}

Current visual question answering (VQA) models struggle with consistency.
They often correctly answer complex reasoning questions, i.e, those requiring common sense knowledge and logic on top of perceptual capabilities, but fail on associated low level perception questions, i.e., those directly related to the visual content in the image.
For e.g., in Fig~\ref{fig:approach}, models answer the reasoning question ``Was this taken in the daytime?'' correctly, but fail on the associated perception question ``Is the sky bright?'' indicating that the models likely answered the reasoning question correctly for the wrong reason(s). 
In this work, we explore the usefulness of leveraging information about~\subqs, i.e., low level perception questions relevant to a reasoning question, and~\irrqs, i.e., any other questions about the image that are unrelated to the reasoning question, to improve consistency in VQA models. 


\citet{squint} have studied this problem and introduced the VQA-Introspect dataset that draws a distinction between higher-level reasoning questions and lower-level perception sub-questions. 
We augment this dataset with additional perception questions from the VQAv2 dataset such that each $<$image, reasoning question$>$ pair contains a set of relevant perception questions, which we refer to as \emph{sub-questions} (e.g.,``Is the sky bright?'' in Fig~\ref{fig:approach}) and irrelevant perception questions, which we refer to as \emph{irrelevant questions} (e.g., ``Is the train moving?'' in Fig~\ref{fig:approach}) throughout this paper. 

We use Gradient-based Class Activation Mapping (Grad-CAM) vectors \citep{Selvaraju_2019} -- a faithful function of the model's parameters, question, answer and image -- to propose an interpretability technique that determines the questions most strongly correlated with a reasoning question for a model. 
This is measured by ranking questions based on the cosine similarity of their Grad-CAM vectors with that of the reasoning question. 
We find that even top-performing VQA models often rank irrelevant questions higher than relevant questions. 

Motivated by this, we introduce a new approach based on contrastive gradient learning to fine-tune a VQA model by adding a loss term that enforces relevant~\subqs~to be ranked higher than~\irrqs~while answering a reasoning question. 
This is achieved by forcing the cosine similarity of the reasoning question's Grad-CAM vector with that of a~\subq~to be higher than with that of an~\irrq.
We find that our approach improves the model's consistency, defined as the frequency with which the model correctly answers a~\subq~given that it correctly answers the reasoning question. 

Additionally, we assess the effects of our approach on visual grounding (i.e, does the model look at the right visual regions while answering the question?) by comparing Grad-CAM heatmaps with human attention maps collected in the VQA-HAT dataset \citep{vqahat}. 
We find that our approach of enforcing this language-based alignment through better ranking of~\subqs~also improves visual grounding. 

\section{Related Work}

\label{related_work}
\noindent\textbf{Visual Question Answering}. 
The VQA task \citep{vqa} requires answering a free form natural language question about visual content in an image. 
Previous work has shown that models often do well on the task by exploiting language and dataset biases~\cite{assume, zhang2015yin, sain2018overcoming, guo2019quantifying, manjunatha2018explicit}. 
In order to evaluate the consistency of these models,~\citet{squint} 
 collected a new dataset, VQA-Introspect, containing human explanations in the form of sub-questions and answers for reasoning questions in the VQA dataset. 

\noindent\textbf{Model Interpretability}. 
While prior work has attempted to explain VQA decisions in the visual modality \cite{Selvaraju_2019, hint, qiao2017exploring}, the multi-modal task of VQA has a language component which cannot always be explained visually, i.e., visual regions can be insufficient to express underlying concepts \cite{goyal2016transparent, Hu_2017_ICCV}. \citet{Liang_2019} introduced a spatio-temporal attention mechanism to interpret a VQA model's decisions across a set of consecutive frames. Our work, operating on a single image, interprets the model's decisions using Grad-CAM vectors which are more faithful to the model's parameters than attention maps. 
\citet{park2018multimodal} and \citet{wu:blackboxnlp19} generate textual justifications through datasets curated with human explanations. 
Our approach using Grad-CAM vectors differs by being fully self-contained and faithful to the model, requiring no additional parameters for interpreting its decisions. 



\noindent\textbf{Aligning network importances}. 
\citet{hint} introduced an approach to align visual explanations with regions deemed important by humans, thereby improving visual grounding in VQA models. 
In followup work, \citet{squint} introduced an approach to align attention maps for the reasoning question and associated perception~\subqs~from VQA-Introspect to improve language based grounding. 
In contrast to attention maps, our work encourages Grad-CAM vectors of a reasoning question to be closer to those of~\subqs~and farther away from those of~\irrqs. Intuitively, this means that we are making the neurons used while answering a reasoning question to be similar to those used while answering a~\subq~and dissimilar to those used while answering an~\irrq.
Our experiments show that this alignment improves the model's consistency and visual grounding. 

\section{Approach}

\subsection{Preliminaries}

\noindent\textbf{Grad-CAM}.
Grad-CAM was introduced by \citet{Selvaraju_2019} as a technique to obtain visual explanations from any CNN-based deep neural network. 
In this work, we adopt Grad-CAM to compute the contribution of a neuron at the layer in a VQA model where the vision and language modalities are combined. 
This is computed by taking the gradient of the predicted output class score with respect to the neuron activations in the layer. 
We then point-wise multiply this with the corresponding activations to obtain our Grad-CAM importance vector. 
Specifically, if $y^c$ denotes the score of the ground-truth output class and $A_{k}$ the activations of layer $k$ of the model, the Grad-CAM importance vector $G^c_{k}$ (or simply, Grad-CAM vector) is computed as follows,
\begin{equation}
\label{eq1}
\begin{split}
G^c_{k} &= \frac{\partial y^c}{\partial A_{k}} * A_{k}
\end{split}
\end{equation}

Unlike Grad-CAM visualizations, these vectors are not visually interpretable as they are not computed on the final convolutional layer of the CNN. 



\noindent\textbf{Consistency in VQA models}. As defined in \citet{squint}, the consistency of a VQA model refers to the proportion of \textit{sub-questions} answered correctly, given that their corresponding reasoning questions were answered correctly. If a model is inconsistent, it is likely relying on incorrect perceptual signals or biases in the dataset to answer questions. Models that are consistent and based on appropriate perceptual signals are more likely to be reliable, interpretable and trustworthy. 

\subsection{Sub-question Oriented Tuning} \label{sec:sort}
The key idea behind Sub-question Oriented Tuning (SOrT) is to encourage the neurons most strongly relied on (as assessed by Grad-CAM vectors) while answering a reasoning question (``Was this taken in the daytime?'' in Fig~\ref{fig:approach}) to be similar to those used while answering the~\subqs~(``Is the sky bright?'') and dissimilar to those used while answering the~\irrqs~(``Is the train moving?''). 
This enforces the model to use the same visual and lingustic concepts while making predictions on the reasoning question and the~\subqs. 
Our loss has the following two components.  


\noindent\textbf{Contrastive Gradient Loss}. 
With the Grad-CAM vectors of the reasoning question $(G_{R})$,~\subq~$(G_{S})$ and~\irrq~$(G_{I})$, we formalize our intuition of a contrastive gradient loss $\mathcal{L}_{\text{CG}}$ as,

\begin{equation}
\label{eq1}
\begin{split}
\mathcal{L}_{\text{CG}} &= \text{max}\left(0,\overbrace{\frac{G_{R} \cdot G_{I}}{|G_{R}||G_{I}|}}^{\text{cosine-sim($G_{R}$, $G_{I}$)}} - \underbrace{\frac{G_{R} \cdot G_{S}}{|G_{R}||G_{S}|}}_{\text{cosine-sim($G_{R}$, $G_{S}$)}}\right)
\end{split}
\end{equation}

\noindent\textbf{Binary Cross Entropy Loss}. To retain performance of the model on the base task of answering questions correctly, we add a Binary Cross Entropy Loss term ($\mathcal{L}_{\text{BCE}}$) that penalizes incorrect answers for all the questions.

\noindent\textbf{Total Loss}. 
Let $\text{o}_{\text{R}}$, $\text{gt}_{\text{R}}$, $\text{o}_{\text{S}}$, $\text{gt}_{\text{S}}$, $\text{o}_{\text{I}}$ and $\text{gt}_{\text{I}}$ represent the predicted and ground-truth answers for the reasoning,~\subqs~and~\irrqs~respectively, and $\lambda_{1}$, $\lambda_{2}$, $\lambda_{3}$ be tunable hyper-parameters. 
Our total loss $\mathcal{L}_{\text{SOrT}}$ is,
\begin{equation}
\label{eq1}
\begin{split}
     \mathcal{L}_{\text{SOrT}} & = \mathcal{L}_{\text{CG}} + \lambda_{1} \mathcal{L}_{\text{BCE}}(\text{o}_{\text{R}}, \text{gt}_{\text{R}}) \\
                        & + \lambda_{2} \mathcal{L}_{\text{BCE}}(\text{o}_{\text{S}}, \text{gt}_{\text{S}}) + \lambda_{3} \mathcal{L}_{\text{BCE}}(\text{o}_{\text{I}}, \text{gt}_{\text{I}})
\end{split}
\end{equation}


\section{Experiments} \label{sec:experiments}
\begin{table*}[t] \footnotesize
\renewcommand*{\arraystretch}{1.3}
\setlength{\tabcolsep}{6pt}
\begin{center}
\resizebox{2\columnwidth}{!}{
\begin{tabular}{l l c  c c c c c c  c c c c c c c c}
\hline
& & & \multicolumn{4}{c}{Consistency Metrics} & & \multicolumn{2}{c}{Accuracy Metrics} &  \multicolumn{4}{c}{Ranking Metrics}\\
& Method & & \colorbox{lightgreen}{R\cmark~S\cmark} $\textcolor{green}{\uparrow}$ & \colorbox{lightblue}{R\cmark~S\xmark} $\textcolor{red}{\downarrow}$ & \colorbox{lightyellow}{R\xmark~S\cmark} $\textcolor{red}{\downarrow}$ & \colorbox{lightpurple}{R\xmark~S\xmark} $\textcolor{red}{\downarrow}$ & Consistency\% $\textcolor{green}{\uparrow}$ & Reas. Accuracy\% $\textcolor{green}{\uparrow}$ & VQA Accuracy\% $\textcolor{green}{\uparrow}$& MP@1 $\textcolor{green}{\uparrow}$ & Ranking Accuracy $\textcolor{green}{\uparrow}$ & & MRR $\textcolor{green}{\uparrow}$ & WPR $\textcolor{red}{\downarrow}$ \\
\hline
& Pythia && 50.61 & 19.88 & \textbf{17.15} & 12.36 & 71.81 & \textbf{75.15} & \textbf{64.95} & 57.75 & 30.33 & & 71.87 & 52.75 \\
& Pythia + SQuINT  && 53.03 & 17.58 & 18.63 & 10.74 & 75.10 & 74.95 & 64.75 & 55.87 & 29.45 & & 71.49 & \textbf{39.20} \\

& Pythia + SOrT (only SQ) && 54.34 & 15.44 & 20.08 & 10.14 & 77.87 & 73.78 & 63.69 & 59.47 & 30.73 & & 74.22 & 41.06 \\
& \textbf{Pythia + SOrT (SQ + IQ)} && \textbf{54.62} & \textbf{15.09} & 20.31 & \textbf{9.97} & \textbf{78.35} & 74.18 & 64.07 & \textbf{61.73} & \textbf{31.90} & & \textbf{74.43} & 40.03 \\
\hline
\end{tabular}}\\[5pt]
\vspace{-10pt}
\caption{Results on the Consistency, Accuracy and Ranking metrics described in Sec~\ref{sec:metrics}. Consistency and Ranking are benchmarked on the val split of VQA-Introspect, while Reasoning Accuracy and VQA Accuracy are on the reasoning and val splits of VQAv2 respectively. SQ refers to~\subqs~and IQ to~\irrqs.}
\label{tab:results}
\end{center}
\vspace{-10pt}
\end{table*}
\noindent \textbf{Dataset.} Our dataset pools VQA-Introspect and VQAv2 such that for every reasoning question in VQA-Introspect, we have a set of $<$\subq, \emph{answer}$>$ pairs and a set of $<$\irrq, \emph{answer}$>$ pairs. The training/val splits contain 54,345/20,256 $<$image, reasoning question$>$ pairs with an average of 2.58/2.81~\subqs~and 7.63/5.80~\irrqs~for each pair. 

\noindent \textbf{Baselines.}
We compare SOrT against the following baselines:
\noindent 1) \textbf{Pythia}.
\noindent 2) \textbf{SQuINT} in which, as discussed in Sec~\ref{related_work},~\citet{squint} fine-tuned Pythia with an attention alignment loss to ensure that the model looks at the same regions when answering the reasoning and~\subqs. \noindent 3) \textbf{SOrT with only \textit{sub-questions}} in which we discard the \textit{irrelevant questions} associated with a reasoning question and just align the Grad-CAM vectors of the \textit{sub-questions} with that of the reasoning question. This ablation benchmarks the usefulness of the \textit{contrastive} nature of our loss function.
\subsection{Metrics} \label{sec:metrics}
\noindent\textbf{Ranking.} 
We use four metrics to assess the capability of a model to correctly rank its~\subqs.


\noindent 1. \textbf{Mean Precision@1 (MP@1)}. Proportion of $<$image, reasoning question$>$ pairs for which the highest ranked question is a~\subq.

\noindent 2. \textbf{Ranking Accuracy}. Proportion of $<$image, reasoning question$>$ pairs whose~\subqs~are all ranked higher than their~\irrqs~.

\noindent 3. \textbf{Mean Reciprocal Rank (MRR)}. Average value of the highest reciprocal rank of a~\subq~among all the $<$image, reasoning question$>$ pairs. Higher is better.


\noindent 4. \textbf{Weighted Pairwise Rank (WPR) Loss}. For pairs of incorrectly ranked $<$sub, irrelevant$>$ questions, this computes the differences of their similarity scores with the reasoning question. Averaged across all pairs, this computes the \emph{extent} by which rankings are incorrect. Lower is better. \\

\noindent\textbf{Model Performance.} 

\noindent\textbf{1) Quadrant Analysis.}

\noindent a. \colorbox{lightgreen}{\textbf{R\cmark~S\cmark}}
The pairs where reasoning and~\subqs~are both correctly answered.

\noindent b. \colorbox{lightblue}{\textbf{R\cmark~S\xmark}} The pairs where reasoning question is correctly answered, while the~\subq~is incorrectly answered.

\noindent c. \colorbox{lightyellow}{\textbf{R\xmark~S\cmark}} The pairs where reasoning question is incorrectly answered, while the~\subq~is correctly answered.

\noindent d. \colorbox{lightpurple}{\textbf{R\xmark~S\xmark}} The pairs where reasoning and~\subqs~are both incorrectly answered.

\noindent\textbf{2) Consistency.} The frequency with which a model correctly answers a~\subq~given that it correctly answers the reasoning question.
\begin{equation}
\label{eq1}
\begin{split}
     \text{Consistency} & = \frac{\colorbox{lightgreen}{\textbf{
     R\cmark~S\cmark}}}{\colorbox{lightgreen}{\textbf{
     R\cmark~S\cmark}} + \colorbox{lightblue}{\textbf{ R\cmark~S\xmark}}}
\end{split}
\end{equation}
\noindent \textbf{3) Reasoning Accuracy}. The accuracy on the reasoning split of VQAv2 dataset, and

\noindent \textbf{4) Overall Accuracy}. Accuracy on the VQAv2 validation set.

More details on the metrics are in the Appendix.
\subsection{Results} \label{sec:results}
We attempt to answer the following questions: \\
\noindent \textbf{Does SOrT help models better identify the perception questions relevant for answering a reasoning question?}
As described in Sec~\ref{sec:sort}, the model ranks perception questions (\subqs~and~\irrqs) associated with an $<$image, reasoning question$>$ pair according to the cosine similarities of their Grad-CAM vectors with that of the reasoning question. 
As seen in Table~\ref{tab:results}, we find that our approach outperforms its baselines on nearly all the ranking metrics. 
We observe gains of 4-6\% points on MP@1 and MRR, and 1.5-2.5\% points on Ranking Accuracy. 
Likewise, the improvement in WPR - the soft metric that computes the extent by which rankings are incorrect - is a substantial 12\% points over Pythia. 
This confirms that our approach helps better distinguish between the relevant and irrelevant perceptual concepts needed for answering a reasoning question.

\begin{figure}[t!]
\includegraphics[scale=0.235]{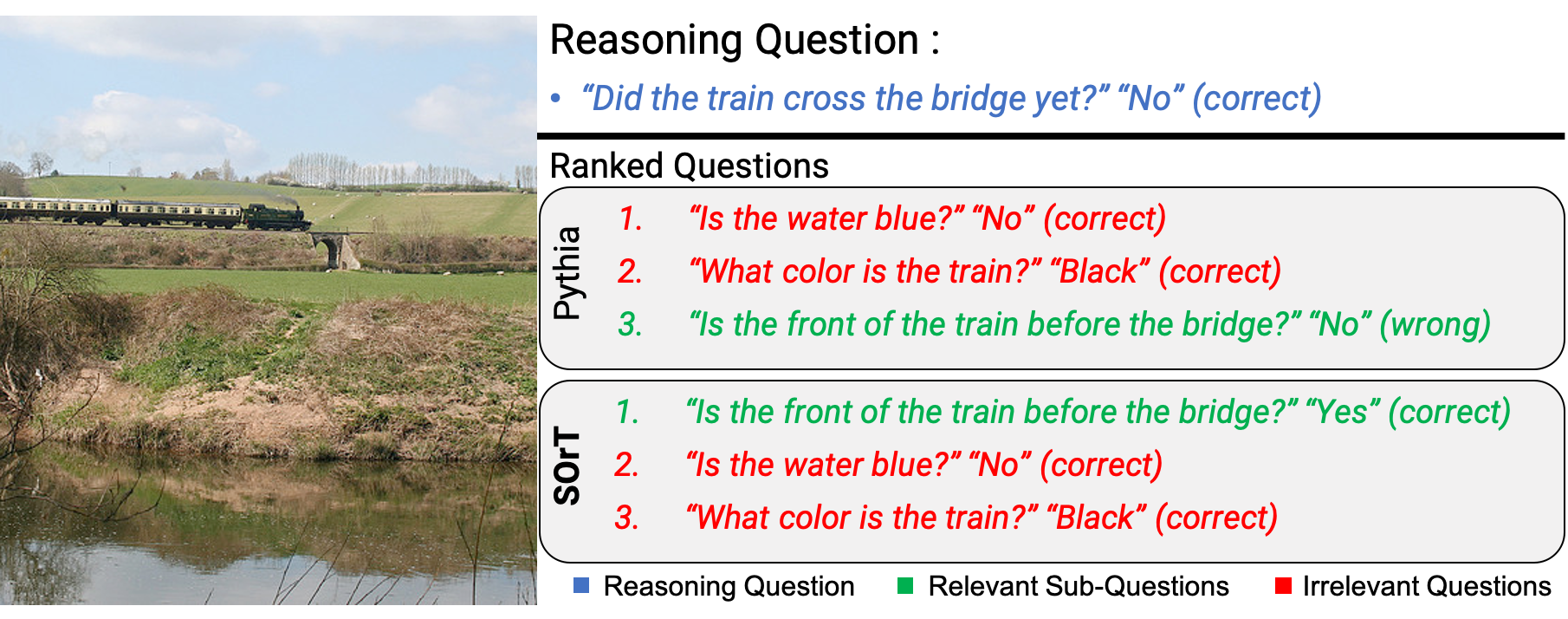}
\caption{An example of improvement in consistency between Pythia (top) and SOrT (below) brought about by better sub-question ranking.}
\label{fig:qual}
\end{figure}

\begin{figure*}[t!]
\includegraphics[scale=0.235]{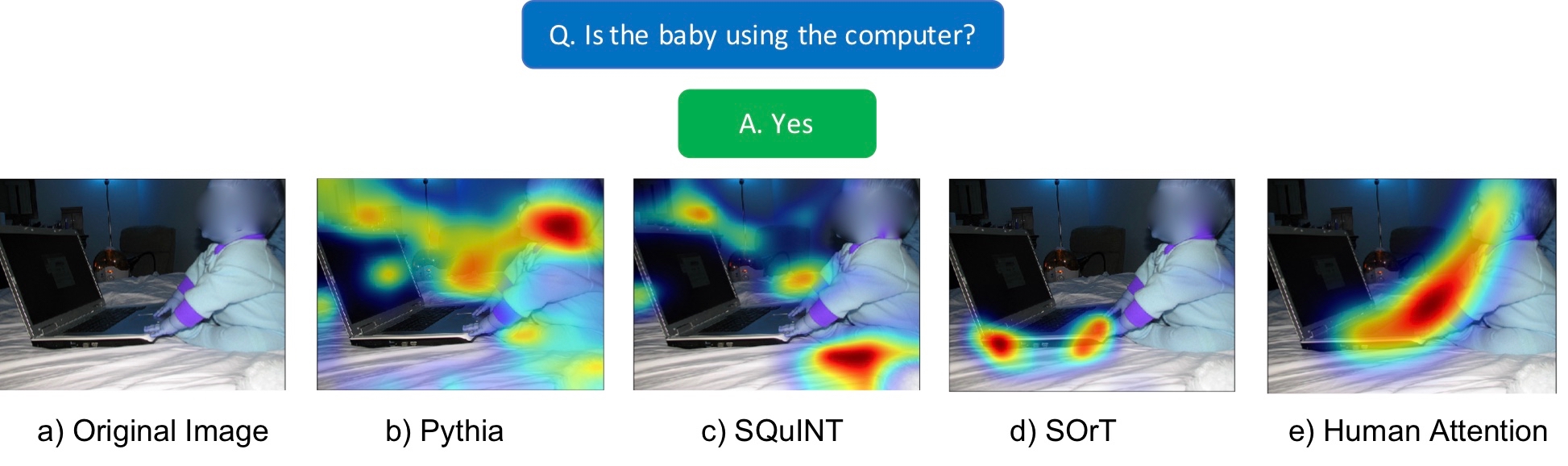}
\caption{A qualitative example of the improvement in visual grounding by SOrT. For the $<$question, answer$>$ pair of $<$\textit{Is the baby using the computer?, Yes}$>$, we see the comparison of the Grad-CAM heatmaps generated by the 3 models and the human attention map. SOrT's heatmap is most closely aligned with that of the human attention map.}
\label{fig:baby}
\end{figure*}

\noindent \textbf{Does recognizing relevant sub-questions make models more consistent?} 
We find that the improved ranking of~\subqs~through SOrT improves consistency by 6.5\% points over Pythia, 3.25\% points over SQuINT and 0.5\% points over an approach that just uses~\subqs~while discarding~\irrqs\footnote{These numbers are averaged values from 10-fold cross validation runs on the val split. The std dev values observed were 0.3 for Pythia and 0.41 for SQuINT and SOrT.}. 
As seen in Table~\ref{tab:results}, the consistency gains are due to significant improvements in the \colorbox{lightgreen}{R\cmark~S\cmark} and \colorbox{lightblue}{R\cmark~S\xmark} quadrants. As seen in Table~\ref{tab:results}, the consistency gains are due to significant improvements in the \colorbox{lightgreen}{R\cmark~S\cmark} and \colorbox{lightblue}{R\cmark~S\xmark} quadrants. 
This comes at the expense of a drop in overall accuracy and reasoning accuracy by $\sim$1\% point, likely due to the active disincentization of memorizing language priors and dataset biases through our contrastive gradient learning approach. 

Gradient-based explanations have been shown to be more faithful to model decisions compared to attention maps~\cite{hint}. 
Our results confirm this by showing that aligning Grad-CAM vectors for reasoning and~\subqs~makes models more consistent compared to SQuINT, which aligns their attention maps. Fig~\ref{fig:qual} shows an example of improved consistency using SOrT. 
The Pythia model answers its~\subq~incorrectly. 
Our approach ranks the relevant~\subq~higher than the \emph{irrelevant} ones and answers it correctly -- thus improving consistency.

These results illustrate a common trade-off among multiple concurrently desirable characteristics of VQA models: accuracy, consistency, interpretability. Building models that maximize all these characteristics presents a challenging problem for future work.

\noindent \textbf{Does enforcing language-based alignment lead to better visual grounding?} 
To evaluate this, we compute visual grounding through Grad-CAM applied on the final convolutional layer.
We then compute the correlation of Grad-CAM heatmaps with the validation split of the VQA-Human ATtenion (VQA-HAT) dataset \cite{vqahat}, comprising 4,122 attention maps. This dataset contains human-annotated `ground truth' attention maps which indicate the regions humans chose to look at while answering questions about images in the VQAv1 dataset. The proposed method to compare human and model-based attention maps in this work was to rank their pixels according to their spatial attention, and then compute the correlation between these two ranked lists.

We find that our approach gets a Spearman rank correlation of $0.103 \pm 0.008$, versus $0.080 \pm 0.009$ for Pythia and $0.060 \pm 0.008$ for SQuINT. 
These statistically significant improvements indicate that enforcing language-based alignment during training improves visual grounding on an unseen dataset. 

A qualitative example that demonstrates the superior visual grounding of SOrT compared to its baselines is shown in Fig \ref{fig:baby}. For the question \emph{Is the baby using the computer?} and its corresponding answer \emph{Yes}, we see that the Grad-CAM heatmap generated by SOrT is closest to that of the human attention map. It is also the only heatmap in this example that actually points to the fingers of the child, which is the essential visual component for answering the question.


\section{Discussion}

In this work, we seek to improve consistency in VQA. We first develop language-based interpretability metrics that measure the relevance of a lower-level perception question while answering a higher-level reasoning question. 
Evaluating state-of-the-art VQA models on these metrics reveals that these models often rank irrelevant questions higher than relevant ones. 
To remedy this, we present SOrT (Sub-question Oriented Tuning), a contrastive gradient learning based approach for teaching VQA models to distinguish between relevant and irrelevant perceptual concepts while answering a reasoning question. SOrT aligns Grad-CAM vectors of reasoning questions with those of~\subqs, while distancing them from those of~\irrqs. We demonstrate SOrT's effectiveness at making VQA models more consistent without majorly affecting accuracy, while also improving visual grounding.


\section{Acknowledgements}
The Georgia Tech effort was supported in part by NSF, AFRL, DARPA, ONR YIPs, ARO PECASE, Amazon. The views and conclusions contained herein are those of the authors and should not be interpreted as necessarily representing the official policies or endorsements, either expressed or implied, of the U.S. Government, or any sponsor.

\bibliography{anthology,emnlp2020}
\bibliographystyle{acl_natbib}
\clearpage

\appendix
\section{Appendices}
\subsection{Experimental Details}

\subsubsection{Algorithms}
We use the Pythia model 
for our experiments. Specifically, for our SOrT approach, we compute Grad-CAM vectors for the reasoning question, \subqs~and~\irrqs~on each image at the layer where the vision and language modalities are combined. We then use customized losses described in Sec~\ref{sec:sort} of the paper. The mathematical computation of consistency is described in Sec~\ref{sec:metrics}, while the ranking metrics are described below.

\noindent\textbf{Mean Precision@1 (MP@1)}. For a given ordering of related questions (based on 1 of the 3 similarity scores), we compute the fraction of pairs in which a relevant perception sub-question was ranked the highest, i.e, had the highest similarity score with that of the reasoning question. This is equivalent to setting a bare-bones expectation of reasoning ability for the model - ``Among all the related questions for a pair, was atleast the highest ranked related question a relevant perception sub-question?"

This is illustrated in an example below across two sets. 

Example Query 1 : ``What is the capital of the USA?" 

Predicted Ranking 1 : [``New York", ``Washington DC", ``San Francisco"]

Ground Truth Answers 1 : [0, 1, 0]

Example Query 2 : ``Where is the Golden Gate Bridge located?"

Predicted Ranking 2 : [``San Francisco", ``Atlanta", ``Los Angeles"]

Ground Truth Answers 2 : [1, 0, 0]

Across these two examples, the Mean Precision@1 value would be $\frac{1}{2}$ since only one of them has its highest ranked item as a correct answer.

\noindent\textbf{Ranking Accuracy}. This computes the proportion of pairs in which all the relevant perception sub-questions are ranked higher than the irrelevant questions. This would represent a perfect ranking capability of the model.

Example Query 1 : ``Cities in Asia."

Predicted Ranking 1 : [``Stockholm", ``Beijing", ``New Delhi"]

Ground Truth Answers 1 : [0, 1, 1]

Example Query 2 : ``Planets in the solar system."

Predicted Ranking 2 : [``Neptune", ``Jupiter", ``Phobos"]

Ground Truth Answers 2 : [1, 1, 0]

The combined Ranking Accuracy across these two examples would be $\frac{1}{2}$ since all the correct answers are ranked higher than the incorrect ones only in the second set. 

\noindent\textbf{Mean Reciprocal Rate (MRR)}. This is a variation of MP@1 which captures the highest rank of a relevant item in a list. In our case, the reciprocal rank is concerned with the highest rank of a relevant perception sub-question among all the ranked related questions for a pair. The reciprocal of this highest relevant rank is averaged across the entire dataset. This is represented in the example below.

Example Query 1 : ``What is the capital of the USA?" 

Predicted Ranking 1 : [``New York", ``Washington DC", ``San Francisco"]

Ground Truth Answers 1 : [0, 1, 0]

Example Query 2 : ``Where is the Golden Gate Bridge located?"

Predicted Ranking 2 : [``San Francisco", ``Atlanta", ``Los Angeles"]

Ground Truth Answers 2 : [1, 0, 0]

Across these two examples, the MRR could be calculated as follows :

\begin{equation}
\label{eq1}
\begin{split}
RR_1 &= \frac{1}{2} \\
RR_2 &= \frac{1}{1} = 1. \\
MRR &= \frac{1}{2} * (RR_1 + RR_2) = \frac{1}{2} * \frac{3}{2} = \frac{3}{4}
\end{split}
\end{equation}

\noindent\textbf{Weight Pairwise Rank (WPR) Loss}. All the above metrics only account for the ranking of the candidate questions for a given pair, but do not consider the \emph{extent} by which these questions differ in their rankings. Concretely, to have a comprehensive understanding of the relevance of each question, we need to account for the magnitude of their similarity scores with the reasoning question in our overall metric. 

For a pair, we create a parallel list of ranked questions in which all the relevant perception sub-questions are higher than the other questions, while retaining the same similarity scores as computed for the originally ranked list. We then compare these two lists pair-wise, i.e, in each index, and sum up the differences of the similarity scores if the rankings are different between the two lists. This provides us a way to measure not just the deviation from the desired order of rankings but also the magnitude of the differences in similarity scores which are responsible for the erroneous rankings. If $S$ could be represented as the set of size $n$ containing all such incorrectly ranked pairs $(r. r')$ with scores ($\alpha$, $\alpha'$), we could compute the WPR loss for each set as the sum of the absolute values of the differences between each $\alpha$ and $\alpha'$. 
\begin{equation}
\label{eq2}
\begin{split}
\text{WPR} &= \frac{\sum_{(r, r') \in S} \left\lvert{\alpha - \alpha'}\right\rvert}{n}
\end{split}
\end{equation}

This is then averaged across the entire dataset.

We illustrate an example for a single set.

Query : "Which of these is a national capital?"

Predicted Ranking With Scores : [(``Mexico City", 0.9), (``Miami", 0.8), (``Copenhagen", 0.7)]

Ground Truth Answers : [1, 0, 1]

Parallel List w.r.t Ground Truth Answers : [(``Mexico City", 0.9), (``Copenhagen", 0.7), (``Miami", 0.8)]
\begin{equation}
\label{eq3}
\begin{split}
\text{WPR} &= \frac{1}{2} * (0.1 + 0.1) = 0.1
\end{split}
\end{equation}

\subsubsection{Source Code}

Our source code is publicly accessible at \href{https://github.com/sameerdharur/sorting-vqa}{https://github.com/sameerdharur/sorting-vqa}.

\subsubsection{Computing Infrastructure}

The computing infrastructure used for training and running the models described in the paper was 1 NVIDIA TITAN Xp GPU.

\subsubsection{Runtime}

The average training time for the model on each combination of hyperparameters was roughly 12 hours.

\subsubsection{Parameters}

The details on the parameters of the model can be found in the Pythia paper referenced in the main section.

\subsubsection{Validation Performance}

The results of the validation performance on each of the different metrics have been included in Sec~\ref{sec:results} of the main section. The metrics have been explained above, with the source code linked above.

\subsubsection{Hyperparameter Search}

For the best performing models, the values of $\lambda$ described in the losses of Sec~\ref{sec:sort} are $\lambda_{1} = \lambda_{2} = 2.27, \lambda_{3} = 0.0003$. These values were selected based on the differing scales of the loss components and chosen from running hyperparameter sweeps. The rest of the hyperparameters were unchanged from those reported for the best performing Pythia model.

A total of $294$ hyperparameter trial runs were conducted with $\lambda_{1}$ and $\lambda_{2}$ ranging from 0.025 to 25, and $\lambda_{3}$ ranging from $1e$-$5$ to $100$.

These values were picked by a combination of uniform sampling and random tuning, and were optimized on a combination of consistency and accuracy. As mentioned in Sec~\ref{sec:results}, the expected validation results fall within the statistical range of the results defined by a standard deviation of 0.3 and 0.41 for Pythia and SQuINT/SOrT. 

\subsubsection{Datasets}

As detailed in Sec~\ref{sec:experiments}, our dataset is a combination of the VQA-Introspect and VQAv2 datasets. In total, our train/val splits contain 54,345/20,256 $<$image, reasoning question$>$ pairs with an average of 2.58/2.81 sub-questions and 7.63/5.80 irrelevant questions for each pair respectively. Subsets of this data have been attached in a zip file with this submission to serve as representative examples. The details on the VQA-Introspect and VQAv2 datasets, which are publicly available, can be found in the corresponding papers cited in the main section. The VQA-HAT dataset used in the visual grounding analysis is also publicly accessible at \href{https://computing.ece.vt.edu/~abhshkdz/vqa-hat/}{this} link.

\end{document}